%% file: main.tex
\documentclass[review]{cvpr}

\usepackage{times}
\usepackage{epsfig}
\usepackage{graphicx}
\usepackage{amsmath}
\usepackage{amssymb}
\usepackage{bm}
\usepackage{amstext}
\usepackage{amsfonts}
\usepackage{cite}
\usepackage{multirow}

\usepackage[pagebackref=true,breaklinks=true,colorlinks,bookmarks=false]{hyperref}

\begin{document}

%%%%%%%%% TITLE
\title{DTGAN: Dual Attention Generative Adversarial Networks for Text-to-Image Generation}

\author{First Author\\
Institution1\\
Institution1 address\\
{\tt\small firstauthor@i1.org}
\and
Second Author\\
Institution2\\
First line of institution2 address\\
{\tt\small secondauthor@i2.org}
}

\maketitle

% input Abstract
\input{Section/abstract}
% input Introduction
\input{Section/introduction}
% input RelatedWork
% \input{Section/RelatedWork}
% input Model
\input{Section/model}
% input Experiments
\input{Section/experiment}
% input Conclusion
\input{Section/conclusion}

{\small
\bibliographystyle{ieee_fullname}
\bibliography{egbib}
}

\end{document}

%% file: Section/abstract.tex
%%%%%%%%% ABSTRACT
\begin{abstract}
Most existing text-to-image generation methods adopt a multi-stage modular architecture which has three significant problems: 1) Training multiple networks increases the run time and affects the convergence and stability of the generative model; 2) These approaches ignore the quality of early-stage generator images; 3) Many discriminators need to be trained. To this end, we propose the Dual Attention Generative Adversarial Network (DTGAN) which can synthesize high-quality and semantically consistent images only employing a single generator/discriminator pair. The proposed model introduces channel-aware and pixel-aware attention modules that can guide the generator to focus on text-relevant channels and pixels based on the global sentence vector and to fine-tune original feature maps using attention weights. Also, Conditional Adaptive Instance-Layer Normalization (CAdaILN) is presented to help our attention modules flexibly control the amount of change in shape and texture by the input natural-language description. Furthermore, a new type of visual loss is utilized to enhance the image resolution by ensuring vivid shape and perceptually uniform color distributions of generated images. Experimental results on benchmark datasets demonstrate the superiority of our proposed method compared to the state-of-the-art models with a multi-stage framework. Visualization of the attention maps shows that the channel-aware attention module is able to localize the discriminative regions, while the pixel-aware attention module has the ability to capture the globally visual contents for the generation of an image.
\end{abstract}

%% file: Section/introduction.tex
%%%%%%%%% BODY TEXT
\section{Introduction}

Generating high-resolution realistic images conditioned on given text descriptions has become an attractive and challenging task in computer vision (CV) and natural language processing (NLP). It has various potential applications, such as art generation, photo-editing and video games. Recent work has achieved crucial improvements in the quality of generated samples through generative adversarial network (GAN) \cite{reed2016generative,radford2015unsupervised, zhang2018photographic,hong2018inferring}, while also boosting the semantic consistency between generated visually realistic images and given natural-language descriptions.
% \begin{figure*}[t]
%   \begin{minipage}[b]{1.0\linewidth}
%   \centerline{\includegraphics[width=180mm]{img0.pdf}}
%   \end{minipage}
%   \caption{The comparison between the current multi-stage architecture and our model. The multi-stage framework (a) generates final images by training three generators and discriminators. The proposed DTGAN (b) is able to synthesize realistic images only using a single generator/discriminator pair. In (a), $G_{0}$-$G_{2}$ are generators and $D_{0}$-$D_{2}$ are discriminators. In (b), $L_{0}$-$L_{6}$ are the dual-attention layers discussed in Section~\ref{sec:3.1}; $G$ and $D$ are our generator and discriminator, respectively.}
%   \vspace{-0.1in}
%   \label{fig0} %% label for entire figure
% \end{figure*}
\begin{figure}[t]
   \begin{minipage}[b]{1.0\linewidth}
   \centerline{\includegraphics[width=85mm]{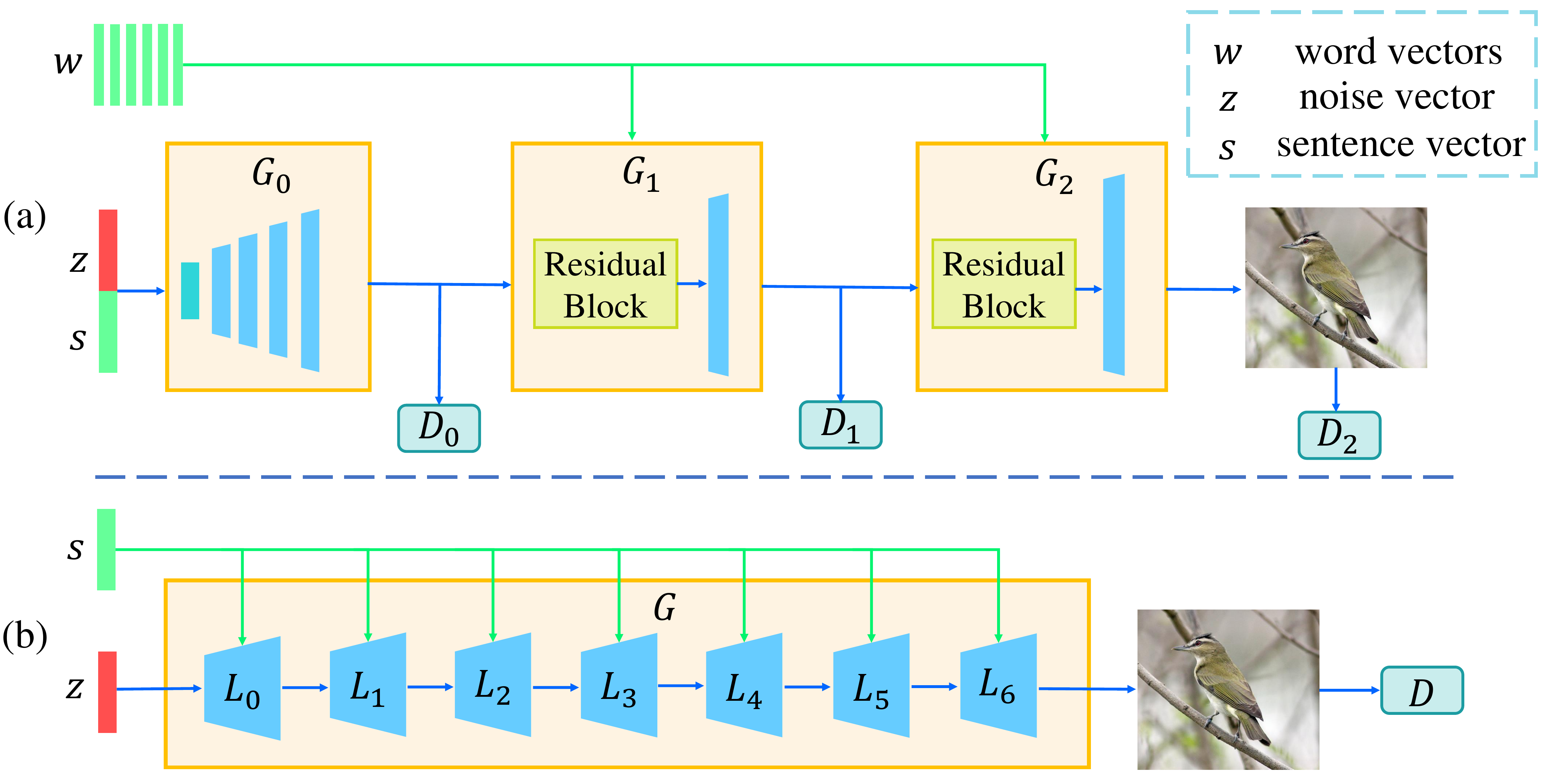}}
   \end{minipage}
   \caption{The comparison between the current multi-stage architecture and our model. The multi-stage framework (a) generates final images by training three generators and discriminators. The proposed DTGAN (b) is able to synthesize realistic images only using a single generator/discriminator pair. In (a), $G_{0}$-$G_{2}$ are generators and $D_{0}$-$D_{2}$ are discriminators. In (b), $L_{0}$-$L_{6}$ are the dual-attention layers discussed in Section~\ref{sec:3}, $G$ and $D$ are our generator and discriminator, respectively.}
   \vspace{-0.1in}
   \label{fig0} %% label for entire figure
\end{figure}

However, most state-of-the-art approaches in text-to-image generation \cite{zhang2017stackgan, zhang2018stackgan++,qiao2019mirrorgan,zhu2019dm,xu2018attngan,li2019controllable,yuan2018text,ma2018gan} are based on a multi-stage modular architecture as shown in Figure~\ref{fig0}(a). Specifically, the network comprises multiple generators which have corresponding discriminators. Furthermore, the generator of the next stage takes the result of the previous stage as the input. This framework has proven to be useful for the task of text-to-image synthesis, but there still exist three significant problems. Firstly, training many networks increases the computation time compared to a unified model and affects the convergence and stability of the generative model \cite{tao2020df}. Even worse, the final generator network cannot be improved if the previous generators do not converge to a global optimum, since the final generator loss does not propagate back. Tao et al. \cite{tao2020df} presents a Matching-Aware zero-centered Gradient Penalty (MA-GP) loss and one-stage framework to overcome the problem of multiple generators and discriminators, but it just utilizes a fully-connected layer to connect feature maps and sentence vector, lacking an efficient mechanism to fuse image features and sentence vector. Secondly, this framework ignores the quality of early-stage generator images which plays a vital role in the resolution of finally-generated images \cite{zhu2019dm}. The generator networks for precursor images are only composed of up-sampling layers and convolution layers, lacking the image integration and refinement process with the input natural-language descriptions. Thirdly, multiple discriminators need to be trained. 

To address the issues mentioned above, we propose a novel Dual Attention Generative Adversarial Network (DTGAN) which can fine-tune the feature maps for each scale according to the given text descriptions, and synthesize high-quality images only using a single generator/discriminator pair. The overall architecture of the DTGAN is illustrated in Figure~\ref{fig0}(b). Our DTGAN consists of four new components, including two new types of attention modules, a new normalization layer, and a new type of visual loss. The first two components in the DTGAN are our designed channel-aware and pixel-aware attention modules which can guide the generator network to focus more on important channels and pixels, and to ignore text-irrelevant channels and pixels by computing attention weights between the global sentence vector and two aforementioned factors. Different from earlier attention models \cite{xu2018attngan,li2019controllable}, we apply the attention scores to fine-tune original feature maps rather than adopt the weighted sum of converted word features as new feature maps. We expect that our proposed attention method will significantly improve the semantic consistency of generated images. In the third ingredient, inspired by Adaptive Layer-Instance Normalization (AdaLIN) \cite{kim2019u}, we present Conditional Adaptive Instance-Layer Normalization (CAdaILN), where the ratio between Instance Normalization \cite{ulyanov2016instance} and Layer Normalization \cite{ba2016layer} is adaptively learned during training and the global sentence vector is employed to scale and shift the normalized result. The CAdaILN function is complementary to the attention modules and helps with controlling the amount of change in shape and texture. As a result, armed with the attention modules and CAdaILN, our network can generate photo-realistic images only exploiting a single generator/discriminator pair. The last proposed component is a new variant for computing visual loss. It is introduced to ensure that generated images and real images have similar color distributions and shape. We expect that the choice of this novel visual loss has a considerable impact on the quality of generated results. 

We perform extensive experiments on the CUB bird \cite{wah2011caltech} and MS COCO \cite{lin2014microsoft} datasets to evaluate the effectiveness of our proposed DTGAN. Both qualitative and quantitative results demonstrate that our approach outperforms existing state-of-the-art models. 
In addition, visualization of the attention maps shows that the channel-aware attention module is able to localize the important parts of an image, while the channel-aware attention module has the ability to capture the globally visual contents. 
The contributions of our work can be summarized as follows:

$\bullet$ To the best of our knowledge, we are the first to propose the fine-tuning on each scale of feature maps using the attention modules and the conditional normalization function, in order to generate high-quality and semantically consistent images only employing a single generator/discriminator pair.

$\bullet$ We design two new types of attention modules to guide the generator to focus on text-relevant channels and pixels, and to refine the feature maps for each scale.

$\bullet$ CAdaILN is presented to help attention modules flexibly control the amount of change in shape and texture.

$\bullet$ We are the first to introduce the visual loss in text-to-image synthesis to enhance the image quality.
\section{Related Work}
\begin{figure*}[t]
  \begin{minipage}[b]{1.0\linewidth}
  \centerline{\includegraphics[width=180mm]{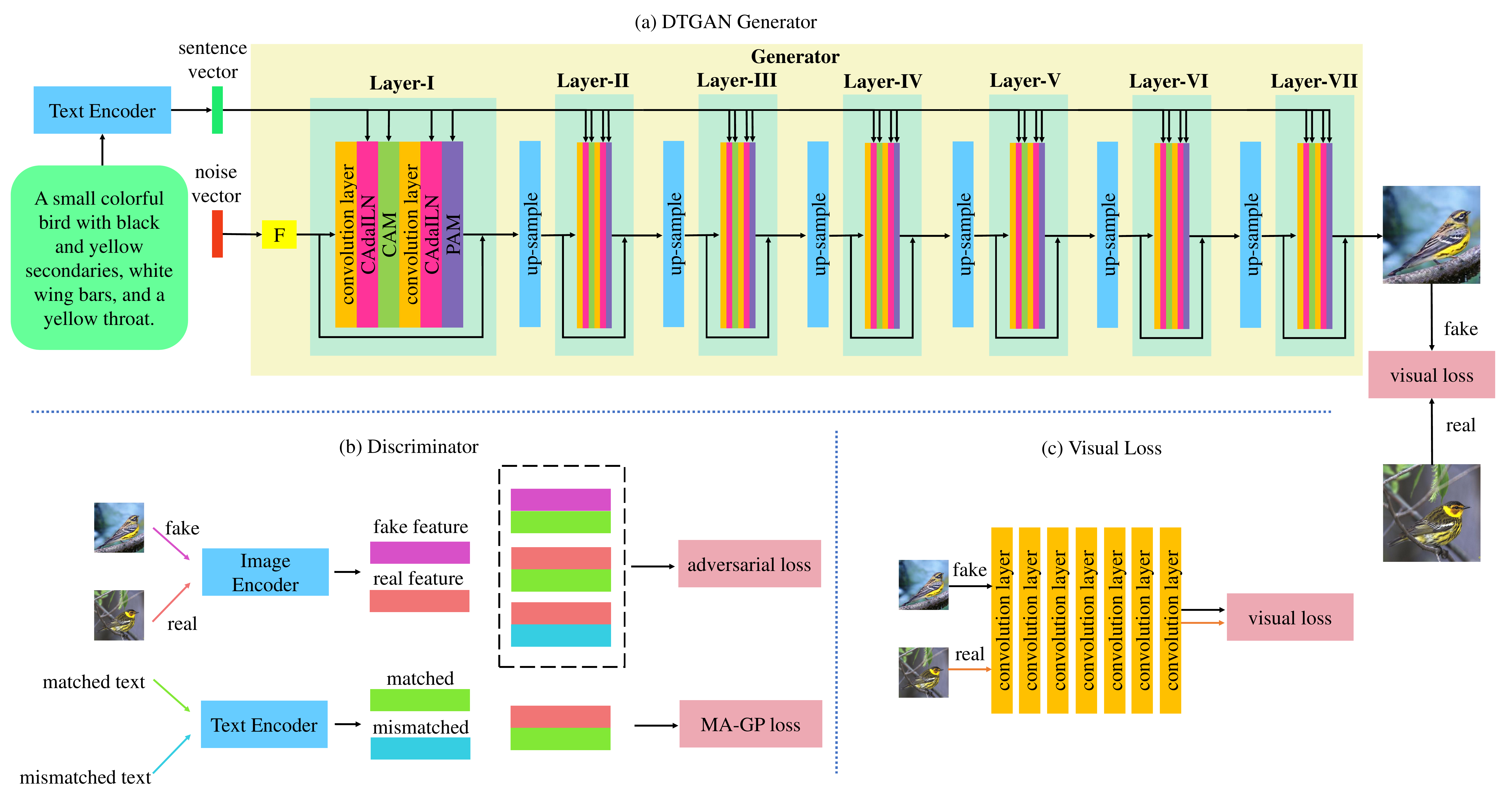}}
  \end{minipage}
  \caption{The architecture of the proposed DTGAN. In (a), F is a fully-connected layer, CAM is a channel-aware attention module discussed in Section~\ref{sec:3.1}, PAM is a pixel-aware attention module discussed in Section~\ref{sec:3.2} and CAdaILN is Conditional Adaptive Instance-Layer Normalization discussed in Section~\ref{sec:3.3}. In (b), MA-GP loss is a Matching-Aware zero-centered Gradient Penalty loss introduced in Section~\ref{sec:3.5}.}
  \vspace{-0.1in}
  \label{fig1} %% label for entire figure
\end{figure*}
\noindent\textbf{Text-to-Image Generation}.
In recent years, the task of text-to-image synthesis has attracted rapidly growing attention from both CV and NLP communities. Thanks to the significant improvements in image generation approaches especially GAN, researchers have achieved inspiring advances in the task of text-to-image generation. 
The conditional GAN \cite{radford2015unsupervised} was first presented by Reed et al. \cite{reed2016generative} to generate plausible images from detailed text descriptions. The problem of text-to-image generation was decomposed by Zhang et al. \cite{zhang2017stackgan, zhang2018stackgan++} into multiple stages. Each stage accomplished the corresponding task by using different generators and discriminators.We aim to generate high-quality images with photo-realistic details just employing a pair of generator and discriminator. Qiao et al. \cite{qiao2019mirrorgan} introduced the image caption model to regenerate the text description from the generated image, in order to enhance the semantic relevancy between the text description and visual content. Zhu et al. \cite{zhu2019dm} applied a dynamic memory module to refine the image quality of the initial stage.

\noindent\textbf{Attention}. 
Attention mechanisms play a vital role in bridging the semantic gap between vision and language. They have been extensively explored in the interdisciplinary fields, such as image captioning \cite{pan2020x, cornia2020meshed}, visual question answering \cite{anderson2018bottom, kim2018bilinear, li2019relation} and visual dialog \cite{niu2019recursive, gan2019multi} . 
Over the past few years, there have been some attention methods for the task of text-to-image generation. Xu et al. \cite{xu2018attngan} utilized a word-level spatial attention mechanism to obtain the relationship between the subregions of the generated image and the words in the input text. The most relevant subregions to the words were very focused. Li et al. \cite{li2019controllable} designed a word-level channel-wise attention mechanism on the basis of Xu et al. \cite{xu2018attngan}, simultaneously taking spatial and channel information into account. However, the aforementioned attention works adopt the weighted sum of converted word features as the new feature map which is largely different from the original feature map. We propose to fine-tune the original feature map using the channel-aware attention weights and the pixel-aware attention weights.

%% file: Section/model.tex
\section{DTGAN for Text-to-Image Generation}
\label{sec:3}
In this section, we elaborate on our proposed DTGAN which is shown in Figure~\ref{fig1}. Unlike prior works \cite{zhang2017stackgan, zhang2018stackgan++,qiao2019mirrorgan,zhu2019dm,xu2018attngan,li2019controllable,yuan2018text,ma2018gan}, our goal is to generate a high-quality and visually realistic image which semantically aligns with a given natural-language description only employing  a single generator/discriminator pair. To this end, we present four significant components: a channel-aware attention module, a pixel-aware attention module, Conditional Adaptive Instance-Layer Normalization (CAdaILN) and a new type of visual loss. Each of them will be discussed in detail after briefly describing the overall framework of our model.  
% \subsection{Architecture}
% \label{sec:3.1}
% \begin{figure*}[t]
%   \begin{minipage}[b]{1.0\linewidth}
%   \centerline{\includegraphics[width=180mm]{img2.pdf}}
%   \end{minipage}
%   \caption{Overview of the proposed channel-aware attention module. GAP and GMP denote global average pooling and global max pooling, respectively.}
%   \vspace{-0.1in}
%   \label{fig2} %% label for entire figure
% \end{figure*}

As shown in Figure~\ref{fig1}, our architecture is composed of a text encoder and a generator/discriminator pair. For text encoder, we adopt a bidirectional Long Short-Term Memory (LSTM) network \cite{schuster1997bidirectional} to learn the semantic representation of a given text description. Specifically, in the bidirectional LSTM layer, two hidden states are employed to capture the semantic meaning of a word and the last hidden states are utilized to represent the sentence features. 
% The word vector is indicated by $w\in R^{D\times T}$ and the sentence vector is denoted by $s\in R^{D}$.
% \begin{figure*}[t]
%   \begin{minipage}[b]{1.0\linewidth}
%   \centerline{\includegraphics[width=180mm]{img3.pdf}}
%   \end{minipage}
%   \caption{Overview of the proposed pixel-aware attention module. SAP and SMP denote average pooling and max pooling in the spatial dimension, respectively.}
%   \vspace{-0.1in}
%   \label{fig3} %% label for entire figure
% \end{figure*}

The generator network of the DTGAN takes a global sentence vector and a noise vector as the input and consists of seven dual-attention layers which are responsible for different scales of feature maps. Each dual-attention layer comprises two convolution layers, two CAdaILN layers, a channel-aware attention module and a pixel-aware attention module.
Mathematically,
\begin{align}
&h_{0}=F_{0}(z) \\
&h_{1}=F_{1}^{Dual}(h_{0},s) \\
&h_{i}=F_{i}^{Dual}(h_{i-1}\uparrow,s)  \quad for \quad i=2,3,...,7 \\
&o=G_{c}(h_{7}) 
\end{align}
where $z$ is a noise vector from the normal distribution, $F_{0}$ is a fully-connected layer, $F_{i}^{Dual}$  is our proposed dual-attention layer, $G_{c}$ is the last convolution layer, $h_{0}$ is the output of the first fully-connected layer, $h_{1}$-$h_{7}$ are the outputs of dual-attention layers and $o$ is the generated image.

In order to take into account both channel information and spatial pixels, we present the channel-aware and pixel-aware attention modules. Different from AttnGAN \cite{xu2018attngan} and ControlGAN \cite{li2019controllable}, we attend to fine-tune original feature maps for each scale using attention modules, rather than adopt the weighted sum of converted word features as the new feature maps. The experiments conducted on benchmark datasets show the superiority of our proposed attention modules compared to AttnGAN and ControlGAN.
\subsection{Channel-aware Attention Module}\label{sec:3.1}
The feature map of each channel at the convolution layer plays different roles in generating the image which semantically aligns with the given text description. Without fine-tuning the channel maps at the generative stage according the text description, the generated result can lack the semantic relevancy to the given text description. Thus, we introduce a channel-aware attention module to guide the generator to focus on text-relevant channels and ignore minor channels.
% \begin{figure}[t]
%   \begin{minipage}[b]{1.0\linewidth}
%   \centerline{\includegraphics[width=85mm]{img2.pdf}}
%   \end{minipage}
%   \caption{Overview of the proposed channel-aware attention module. GAP and GMP denote global average pooling and global max pooling, respectively.}
%   \vspace{-0.1in}
%   \label{fig2} %% label for entire figure
% \end{figure}
\begin{figure}[t]
   \begin{minipage}[b]{1.0\linewidth}
   \centerline{\includegraphics[width=85mm]{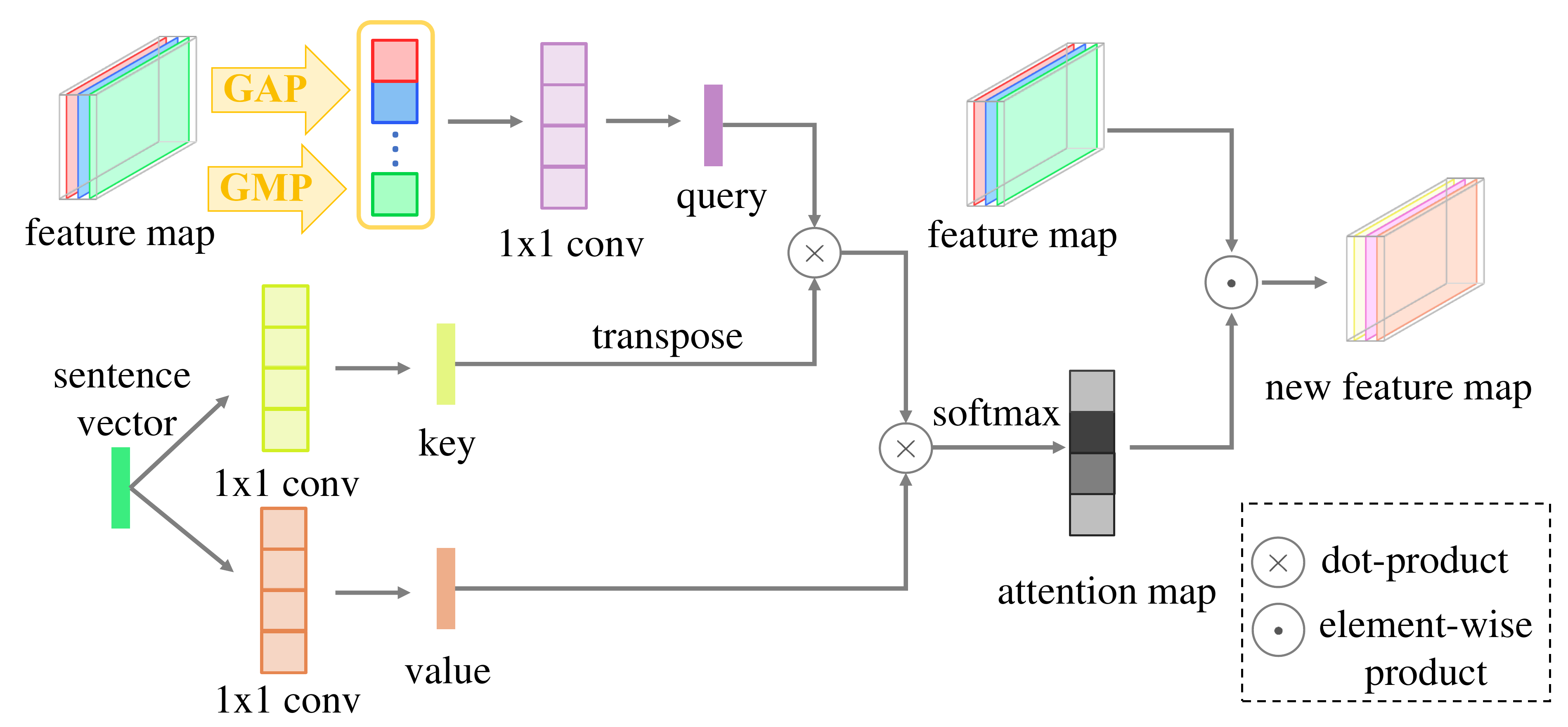}}
   \end{minipage}
   \caption{Overview of the proposed channel-aware attention module. GAP and GMP denote global average pooling and global max pooling, respectively.}
   \vspace{-0.1in}
   \label{fig2} %% label for entire figure
\end{figure}

The process of the channel-aware attention module is shown in Figure~\ref{fig2}. The channel-aware attention module takes two inputs: the feature map $h$ and the global sentence vector $s$ . Firstly, we perform global average pooling and global max pooling on $h$  to extract the channel features: $x_{a}\in R^{C\times 1\times 1}$ and $x_{m}\in R^{C\times 1\times 1}$. Global average pooling is to obtain the information of the whole feature map, while global max pooling focuses on the most discriminative part \cite{zhou2016learning}. Mathematically,
\begin{align}
&x_{a}=\textnormal{GAP}(h)  \\
&x_{m}=\textnormal{GMP}(h)
\end{align}
where $\textnormal{GAP}$ denotes global average pooling, $\textnormal{GMP}$ is global max pooling.

Then, we adopt a query, key and value setting to capture the semantic relevancy between channels and the input text, where $x_{a}$ and $x_{m}$ are used as the query and $s$ is selected as the key and the value. It is defined as:
\begin{align}
&q_{ac}=W_{qa}x_{a},q_{mc}=W_{qm}x_{m}  \\
&k_{c}=W_{kc}s,v_{c}=W_{vc}s 
\end{align}
where $W_{qa}$, $W_{qm}$, $W_{kc}$ and $W_{vc}$ are the projection matrixes which are implemented as 1$\times$1 convolutions.

Assuming that the dot products between  the sentence-level key $k_{c}^{T}$ and the average-pooling query $q_{ac}$, the max-pooling query $q_{mc}$ can capture meaningful features, the attention scores of channel maps are achieved through the following attention mechanism \cite{vaswani2017attention}:
\begin{align}
&\tilde{\alpha}_{a}^{c}=q_{ac}\cdot k_{c}^{T},\tilde{\alpha}_{m}^{c}=q_{mc}\cdot k_{c}^{T}  \\
&\alpha_{a}^{c}=\textnormal{softmax}(\tilde{\alpha}_{a}^{c}\cdot v_{c})   \\
&\alpha_{m}^{c}=\textnormal{softmax}(\tilde{\alpha}_{m}^{c}\cdot v_{c}) 
\end{align}
where $\tilde{\alpha}_{a}^{c}$ and $\tilde{\alpha}_{m}^{c}$ represent the semantic similarity between channel maps and the global sentence vector, $\alpha_{a}^{c}\in R^{C\times 1\times 1}$ and $\alpha_{m}^{c}\in R^{C\times 1\times 1}$  denote the final attention weights of channels for global average pooling and global max pooling, respectively, $\tilde{\alpha}_{a}^{c}$, $\tilde{\alpha}_{m}^{c}$, $\alpha_{a}^{c}$ and $\alpha_{m}^{c}$  are all computed by dot products.

After acquiring the attention weights of channels, we multiply them and the original feature maps to update the feature maps. It is denoted as:
\begin{align}
&o_{ac}=\alpha_{a}^{c}\odot h  \\
&o_{mc}=\alpha_{m}^{c}\odot h 
\end{align}
where $\odot$ is the element-wise multiplication. By doing so, the network will focus on the channels which are more semantically related to the given text description.

Meanwhile, the results of  global average pooling and global max pooling are fused through concatenation. Specifically,
\begin{equation}
o_{c}=\sigma(W_{c}[o_{ac};o_{mc}]) 
\end{equation}
where $W_{c}$ is implemented as 1$\times$1 convolution, $\sigma$ is a nonlinear function, such as ReLU. 

We further apply an adaptive residual connection \cite{zhang2019self} to generate the final result. It is defined as follows:
\begin{equation}
y_{c}=\gamma _{c}*o_{c}+h
\end{equation}
where $\gamma _{c}$ is a learnable parameter which is initialized as 0. 

As can be seen from above, our designed channel-aware attention model is a fine-tuning module based on channel information and text features. Moreover, it is applied on each scale of feature maps to improve the semantic consistency of generated samples at the generative stage.
\subsection{Pixel-aware Attention Module}\label{sec:3.2}
An image is composed of correlated pixels which are of central importance for the quality and semantic consistency of synthesized images. Thus, we propose a pixel-aware attention module to effectively model the relationships between spatial pixels and the given natural-language description, and to make the important pixels receive more attentions from the generator.
% \begin{figure}[t]
%   \begin{minipage}[b]{1.0\linewidth}
%   \centerline{\includegraphics[width=85mm]{img3.pdf}}
%   \end{minipage}
%   \caption{Overview of the proposed pixel-aware attention module. SAP and SMP denote average pooling and max pooling in the spatial dimension, respectively.}
%   \vspace{-0.1in}
%   \label{fig3} %% label for entire figure
% \end{figure}
\begin{figure}[t]
  \begin{minipage}[b]{1.0\linewidth}
  \centerline{\includegraphics[width=85mm]{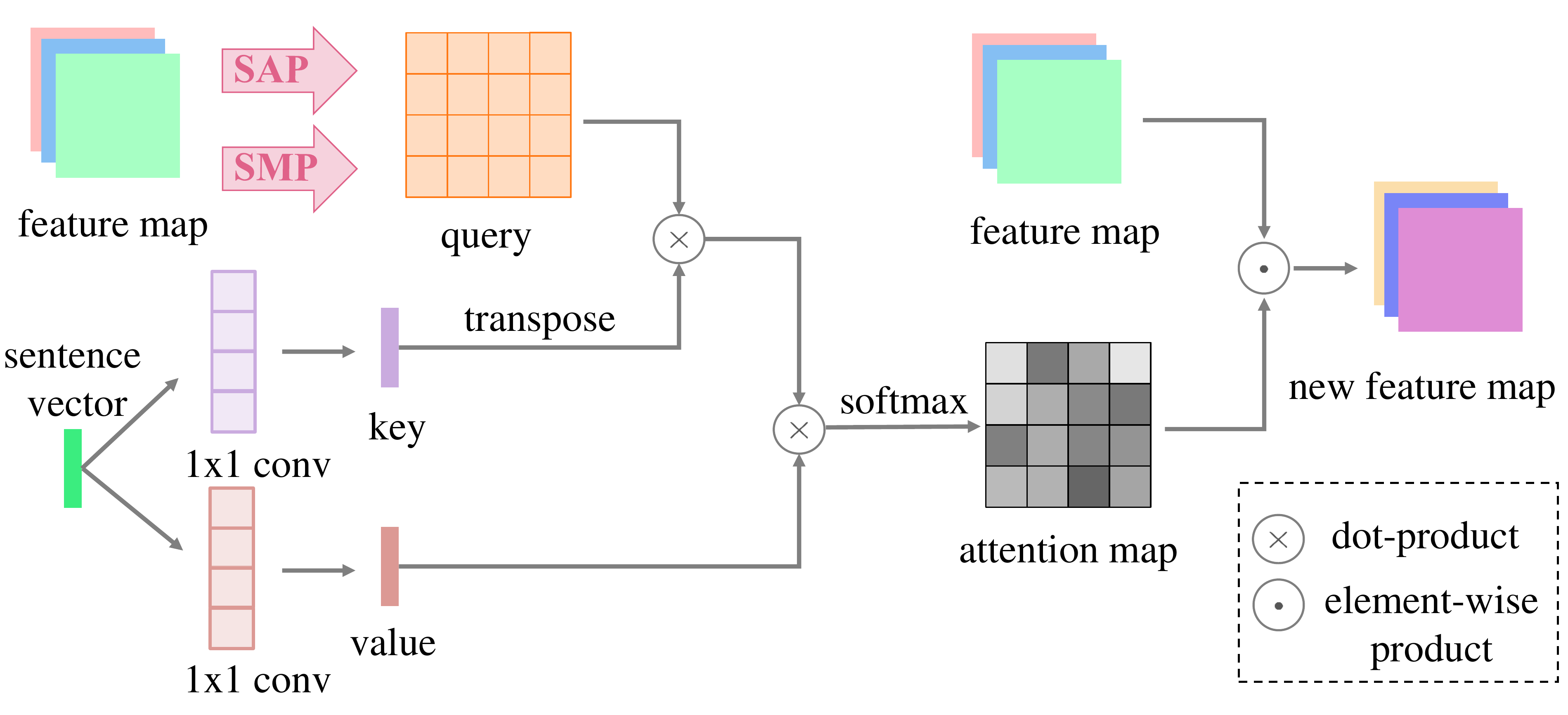}}
  \end{minipage}
  \caption{Overview of the proposed pixel-aware attention module. SAP and SMP denote average pooling and max pooling in the spatial dimension, respectively.}
  \vspace{-0.1in}
  \label{fig3} %% label for entire figure
\end{figure}

The framework of the pixel-aware attention module is illustrated in Figure~\ref{fig3}. Given the feature map $\hat{h}$ and the global sentence vector $s$, we first exploit average pooling and max pooling to process $\hat{h}$. Specifically,
\begin{align}
&e_{a}=\textnormal{SAP}(\hat{h}) \\
&e_{m}=\text{SMP}(\hat{h}) 
\end{align}
where $\textnormal{SAP}$ and $\textnormal{SMP}$ represent average pooling and max pooling in the spatial dimension, respectively. $e_{a}\in R^{1\times H\times W}$ and $e_{m}\in R^{1\times H\times W}$ are the new feature maps.

Then, $s$ is adopted as the key and the value:
\begin{equation}
k_{p}=W_{kp}s,v_{p}=W_{vp}s
\end{equation}
where $W_{kp}$ and $W_{vp}$ are the learnable matrixes which are implemented as 1$\times$1 convolutions.
% where $W_{kp}\in R^{D\times C}$ and $W_{vp}\in R^{D\times C}$ are the learnable matrixes which are implemented as 1$\times$1 convolutions.

After that, we compute the dot products of the new feature maps and the key to get the semantic similarity $\tilde{\alpha}_{a}^{p}$ and $\tilde{\alpha}_{m}^{p}$ between spatial pixels and the global sentence vector. Furthermore, the attention weights are calculated through a softmax function on the dot products of the semantic similarity and the value. It is defined as:
\begin{align}
&\tilde{\alpha}_{a}^{p}=e_{a}\cdot k_{p}^{T},\tilde{\alpha}_{m}^{p}=e_{m}\cdot k_{p}^{T}  \\
&\alpha_{a}^{p}=\text{softmax}(\tilde{\alpha}_{a}^{p}\cdot v_{p}) \\
&\alpha_{m}^{p}=\text{softmax}(\tilde{\alpha}_{m}^{p}\cdot v_{p}) 
\end{align}
where $\alpha_{a}^{p}$ and $\alpha_{m}^{p}$ represent the final attention weights of spatial pixels for average pooling and max pooling, respectively.

Next, same as the channel-aware attention module, we perform a matrix multiplication between the attention weights and the original feature maps to derive the new features $o_{ap}$ and $o_{mp}$: 
\begin{align}
&o_{ap}=\alpha_{a}^{p}\odot \hat{h}  \\
&o_{mp}=\alpha_{m}^{p}\odot \hat{h} 
\end{align}

In addition, we concatenate $o_{ap}$ and $o_{mp}$, and apply a nonlinear function $\sigma$ to compute the result $o_{p}$. Finally, an adaptive residual connection \cite{zhang2019self} is utilized to combine $\hat{h}$ and $o_{p}$. This process is denoted as:
\begin{align}
&o_{p}=\sigma(W_{p}[o_{ap};o_{mp}]) \\
&y_{p}=\gamma_{p}*o_{p}+\hat{h} 
\end{align}
where $W_{p}$ is implemented as 1$\times$1 convolution, $\sigma$ is a nonlinear function, such as ReLU, $\gamma_{p}$ is a learnable parameter which is initialized as 0. 
\subsection{Conditional Adaptive Instance-Layer Normalization (CAdaILN)}\label{sec:3.3}
In order to stabilize the training of GAN \cite{goodfellow2014generative}, most existing text-to-image generation models \cite{xu2018attngan, li2019controllable, zhu2019dm, qiao2019mirrorgan, yin2019semantics} employ Batch Normalization (BN) \cite{ioffe2015batch} which applies the normalization to a whole batch of generated images instead for single ones. However, the convergence of BN heavily depends on the size of a batch \cite{lian2019revisit}. Furthermore, the advantage of BN is not obvious for text-to-image generation since each generated image is more pertinent to the given text description and the feature map itself. To this end, CAdaILN, inspired by U-GAT-IT \cite{kim2019u}, is designed to perform the normalization in the layer and channel on the feature map and its parameters $\gamma$  and $\beta$  are computed by a fully-connected layer from the global sentence vector. CAdaILN is able to help with controlling the amount of change in shape and texture based on the input natural-language text. Mathematically,
\begin{align}
&\hat{a}_{I^{}}=\frac{a-\mu_{I}}{\sqrt{\sigma_{I}^{2}+\epsilon }},\hat{a}_{L^{}}=\frac{a-\mu_{L}}{\sqrt{\sigma_{L}^{2}+\epsilon }} \\
&\gamma=W_{1}s,\beta=W_{2}s   \\
&\hat{a}=\gamma\odot(\rho\odot a\hat{}_{I^{}}+(1-\rho)\odot a\hat{}_{L^{}})+\beta 
\end{align}
where $a$ is the processed feature map, $\mu_{I}$, $\mu_{L}$ and $\sigma_{I}$, $\sigma_{L}$ respectively denote the mean and standard deviation in the channel and layer on the feature map, $\hat{a}_{I^{}}$ and $\hat{a}_{L^{}}$ represent the output of Instance Normalization (IN) and Layer Normalization (LN) respectively, $\gamma$ and $\beta$ are determined by the global sentence vector $s$, $W_{1}$ and $W_{2}$ are fully-connected layers, $\hat{a}$ is the output of CAdaILN. The ratio of IN and LN is dependent on a learnable parameter $\rho$, whose value is constrained to the range of [0, 1]. Moreover, $\rho$ is updated together with generator parameters. 
% \begin{figure}
% \includegraphics[width=\textwidth]{imgs/img2.pdf}
% \caption{A figure caption is always placed below the illustration.
% Please note that short captions are centered, while long ones are
% justified by the macro package automatically.} \label{fig3}
% \end{figure}
\subsection{Visual Loss}
To ensure that generated images and real images have similar color distributions and shape, we propose a new type of visual loss for the generator which is illustrated in Figure~\ref{fig1}. The visual loss plays a vital role in improving the quality and resolution of finally-generated images. It is based on the image features of the real image $I$ and the generated sample $\hat{I}$, and defined as: 
\begin{equation}
L_{vis}=\left | f(I)-f(\hat{I}) \right |_{1}
\end{equation}
where $f(I)$ and $f(\hat{I})$ denote the image features of the real image and the the fake image which are extracted by the discriminator. We impose a L1 loss to minimize the distance between these two image features. To the best of our knowledge, we are the first to present this type of visual loss and apply it in the task of text-to-image generation.
\subsection{Objective Function}\label{sec:3.5}
\noindent\textbf{Adversarial Loss.}
An adversarial loss is employed to match generated samples to the input text. Inspired by \cite{lim2017geometric, tao2020df, zhang2019self}, we utilize the hinge objective \cite{lim2017geometric} for stable training instead of the vanilla GAN objective. The adversarial loss for the discriminator is formulated as:
\begin{equation}
\begin{split}
\mathcal{L}_{\text{adv}}^{D}=&\mathbb{E}_{x\sim p_{\text{data}}}\left [\text{max}(0,1-D(x,s)) \right ]\\
&+\frac{1}{2}\mathbb{E}_{x\sim p_{G}}\left [\text{max}(0,1+D(\hat{x},s)) \right ]\\
&+\frac{1}{2}\mathbb{E}_{x\sim p_{\text{data}}}\left [\text{max}(0,1+D(x,\hat{s})) \right ] 
\end{split}
\end{equation}
where $s$ is a given text description, $\hat{s}$ is a mismatched natural-language description.
%$x$ is the real image from the distribution $p_{data}$, $\hat{x}$ is the generated sample from the distribution $p_{G}$, 

The corresponding generator loss is:
\begin{equation}
\mathcal{L}_{\text{adv}}^{G}=\mathbb{E}_{x\sim p_{G}}\left [D(x,s) \right ]
\end{equation}

\noindent\textbf{Matching-Aware zero-centered Gradient Penalty (MA-GP) Loss.}
To enhance the quality and semantic consistency of generated images, we adopt the MA-GP loss \cite{tao2020df} for the discriminator. The MA-GP loss applies gradient penalty to real images and given text descriptions. It is as follows:
% \begin{flalign*} 
% & 1+1 = 2.  &  
% \end{flalign*}
\begin{equation}
\mathcal{L}_{\text{M}}=\mathbb{E}_{x\sim p_{\text{data}}}\left [(\left \| \nabla_{x}D(x,s) \right \|_{2}+\left \| \nabla_{s}D(x,s) \right \|_{2})^{p}\right ]
\end{equation}

\noindent\textbf{Generator Objective.}
The generator loss comprises an adversarial loss $\mathcal{L}_{\text{adv}}^{G}$ and a visual loss $\mathcal{L}_{\text{vis}}$:
\begin{equation}
\mathcal{L}_{G}=\mathcal{L}_{\text{adv}}^{G}+\lambda_{1} \mathcal{L}_{\text{vis}}
\end{equation}

\noindent\textbf{Discriminator Objective.}
The final objective function of the discriminator is defined as follows:
\begin{equation}
\mathcal{L}_{\text{D}}=\mathcal{L}_{\text{adv}}^{D}+\lambda_{2}\mathcal{L}_{\text{M}}
\end{equation}

%% file: Section/experiment.tex
\section{Experiments}
In this section, we carry out a set of experiments on the CUB bird \cite{wah2011caltech} and MS COCO \cite{lin2014microsoft} datasets, in order to quantitatively and qualitatively evaluate the effectiveness of the proposed DTGAN. The previous state-of-the-art GAN models in text-to-image synthesis, GAN-INT-CLS \cite{reed2016generative}, GAWWN \cite{reed2016learning}, StackGAN++ \cite{zhang2018stackgan++}, AttnGAN \cite{xu2018attngan} and ControlGAN \cite{li2019controllable}, are first compared with our approach. Then, we analyze the significant components of our designed architecture.

\subsection{Datasets}
Two popular datasets in text-to-image generation, CUB bird and MS COCO datasets, are employed to test our method. The CUB dataset encompasses 11,788 images which are split into 8,855 training images and 2,933 test images. The MS COCO dataset contains 123,287 images which are split into 82,783 training images and 40,504 validation images. Each image in the CUB dataset and MS COCO dataset has ten corresponding text descriptions and five corresponding text descriptions, respectively. We preprocess the CUB dataset using the method in StackGAN \cite{zhang2017stackgan}.

\subsection{Evaluation metric}
Inception score (IS) \cite{salimans2016improved}  and Fr\'echet inception distance (FID)  \cite{szegedy2016rethinking} score are extensively employed in the assessment of text-to-image generation. We adopt theses two indexes as the quantitative evaluation measure and generate 30000 images from unseen text descriptions for each metric.

\noindent\textbf{IS.} The IS is to evaluate the visual quality of the generated images via the KL divergence between the conditional class distribution and the marginal class distribution. It’s defined as:
\begin{equation}
I=\text{exp}(\mathbb{E}_{x}[D_{KL}(p(y|\bm{x} )\parallel p(y))])
\end{equation}
where $\bm{x}$ is a generated sample and $y$ is the corresponding label obtained by a pre-trained Inception v3 network \cite{szegedy2016rethinking}. The generated samples are meant to be diverse and meaningful if the IS is large.

\noindent\textbf{FID.} Same as the IS, the FID is also to assess the quality of generated samples by computing the Fr\'echet distance between the generated image distribution and the real image distribution. We use a pre-trained Inception v3 network to achieve the FID. A lower FID means that the generated samples are closer to the corresponding real images. 

However, it is important to note that the IS on the COCO dataset fails to evaluate the image quality and can be saturated, even over-fitted, which is observed by ObjGAN \cite{li2019object} and DFGAN \cite{tao2020df}. Therefore, we do not utilize the IS as the evaluation metric on the COCO dataset. We further find that R-precision \cite{heusel2017gans}, presented by AttnGAN \cite{xu2018attngan}, can not reflect the semantic relation between generated images and given text descriptions, since experimental results show that the R-precision of real images is only 22.22$\%$. Thus, R-precision is not applied on the validation of our model. 

%supplementary materials 
\subsection{Implementation details}
For text encoder, the dimension $D$ is set to 256 and the length of words is set to 18. We implement our model using PyTorch~\cite{paszke2019pytorch}. In the experiments, the network is trained using Adam optimizer~\cite{kingma2014adam} with $\beta_{1}=0.0$ and $\beta_{2}=0.9$. We follow the two timescale update rule (TTUR) \cite{heusel2017gans} and set the learning rate of the generator and the discriminator to 0.0001 and 0.0004. The batch size is set to 24. The hyperparameters $p$ , $\lambda_{1}$ and $\lambda_{2}$ are set to 6, 0.1 and 2, respectively.
\begin{figure*}[t]
   \begin{minipage}[b]{1.0\linewidth}
   \centerline{\includegraphics[width=180mm]{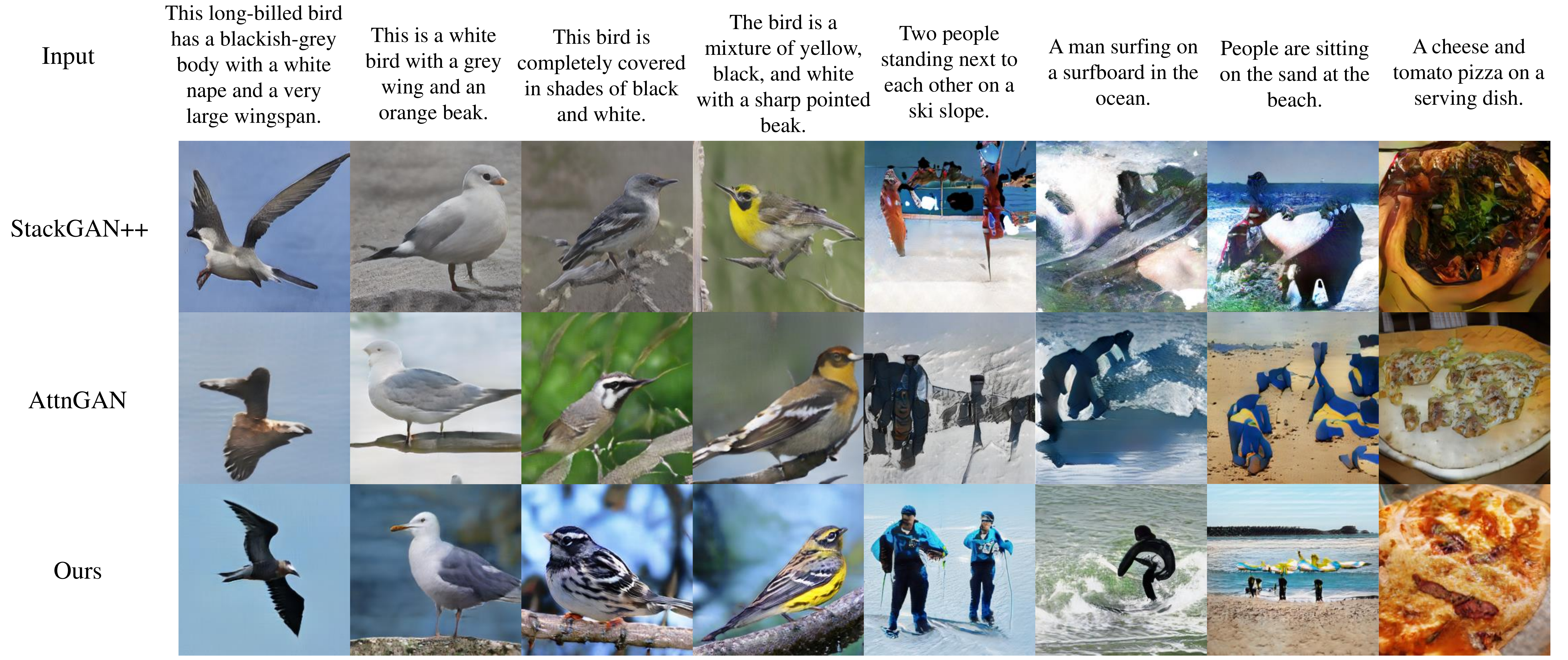}}
   \end{minipage}
   \caption{Qualitative comparison of three approaches conditioned on the text descriptions on the CUB and COCO datasets.}
   \vspace{-0.1in}
  \label{fig6} %% label for entire figure
\end{figure*}
\subsection{Comparison with State of the Art}
\noindent\textbf{Quantitative Results.} We compare our model with prior state-of-the-art GAN approaches in text-to-image synthesis on the CUB and MS COCO datasets. Table~\ref{tab:1} reports the IS of our proposed DTGAN and other compared methods on the CUB dataset. We can observe that our model has the best score, significantly improving the IS from 4.58 to 4.88 on the CUB dataset. The experimental results demonstrate that our DTGAN can generate visually realistic images with higher quality and better diversity than state-of-the-art models.
\begin{table}
\begin{center}
\begin{tabular}{c c}
\hline
Methods & IS $\uparrow$ \\
\hline
GAN-INT-CLS \cite{reed2016generative}& 2.88$\pm$0.04 \\
GAWWN \cite{reed2016learning}& 3.62$\pm$0.07 \\
StackGAN++ \cite{zhang2018stackgan++}& 4.04$\pm$0.05\\
AttnGAN \cite{xu2018attngan}& 4.36$\pm$0.03\\
ControlGAN \cite{li2019controllable}& 4.58$\pm$0.09\\
\hline
Ours &  $\bm{4.88\pm0.03}$\\
\hline
\end{tabular}
\end{center}
\caption{The IS of state-of-the-art approaches and our model on the CUB dataset. The best score is in bold}
\label{tab:1}
\end{table}
\begin{table}
\begin{center}
\scalebox{.97}{
\begin{tabular}{c c c c}
\hline
Datasets & StackGAN++ \cite{zhang2018stackgan++}& AttnGAN  \cite{xu2018attngan}& Ours \\
\hline
CUB &  26.07 & 23.98 & $\bm{16.35}$ \\
COCO &  51.62 & 35.49 & $\bm{23.61}$ \\
\hline
\end{tabular}}
\end{center}
\caption{The FID of StackGAN++, AttnGAN and our model on the CUB and COCO datasets. The best results are in bold.}
\label{tab:2}
\vspace{-0.1in}
\end{table}

The comparison between our method, StackGAN++ \cite{zhang2018stackgan++} and AttnGAN \cite{xu2018attngan} with respect to FID on the CUB and COCO datasets is shown in Table~\ref{tab:2}. We can see that our DTGAN achieves a remarkably lower FID than compared approaches on both datasets, which indicates that our generated data distribution is closer to the real data distribution. Specifically, we impressively reduce the FID from 35.49 to 23.61 on the challenging COCO dataset and from 23.98 to 16.35 on the CUB dataset.
% \begin{figure*}[t]
%   \begin{minipage}[b]{1.0\linewidth}
%   \centerline{\includegraphics[width=180mm]{img6.pdf}}
%   \end{minipage}
%   \caption{Qualitative comparison of three approaches conditioned on the text descriptions on the CUB and COCO datasets.}
%   \vspace{-0.1in}
%   \label{fig6} %% label for entire figure
% \end{figure*}
% \begin{figure*}[t]
%   \begin{minipage}[b]{1.0\linewidth}
%   \centerline{\includegraphics[width=180mm]{img7.pdf}}
%   \end{minipage}
%   \caption{Generated images of the DTGAN by changing the color attributes (in bold) in the text descriptions.}
%   \vspace{-0.1in}
%   \label{fig7} %% label for entire figure
% \end{figure*}

\noindent\textbf{Qualitative Results.} In addition to quantitative experiments, we perform qualitative comparison with StackGAN++ \cite{zhang2018stackgan++} and AttnGAN \cite{xu2018attngan} on both datasets, which is illustrated in Figure~\ref{fig6}. It can be observed that  the details of birds generated by StackGAN++ and AttnGAN are lost ($2^{th}$, $3^{th}$ and $4^{th}$ column), the shape is strange ($1^{th}$, $2^{th}$  and $3^{th}$ column) and the colors are even wrong ($3^{th}$ column). Furthermore, the samples synthesized by these two approaches lack text-relevant objects ($5^{th}$, $6^{th}$ and $7^{th}$ column), the backgrounds are unclear and inconsistent with the given text descriptions ($5^{th}$ and $7^{th}$ column), and the colors are rough ($8^{th}$ column) on the challenging COCO dataset. However, our DTGAN generates more clear and visually plausible images than StackGAN++ and AttnGAN, verifying the superiority of our DTGAN. For instance, as shown in the $1^{th}$ column, owing to the successful application of the visual loss, a long-wingspan bird with vivid shape is produced by the DTGAN, whereas it is too hard for StackGAN++ and AttnGAN to generate this kind of bird. In the meantime, the birds generated by the DTGAN have more details and richer color distributions compared to StackGAN++ and AttnGAN in the $2^{th}$, $3^{th}$ and $4^{th}$ column, since the DTGAN armed with channel-aware and pixel-aware attention modules is able to generate high-resolution images which semantically align with given descriptions. More importantly, our method also yields high-quality and visually realistic results on the challenging COCO dataset. For example, the number of the skiers and surfers is correct, the backgrounds are reasonable and people in the images are clear in the $5^{th}$ and $6^{th}$ column. Moreover, the beach and the sea are very beautiful in the $7^{th}$ column and the pizza looks delicious in the $8^{th}$ column. Generally, these qualitative results confirm the effectiveness of the DTGAN. 
\begin{figure}[t]
  \begin{minipage}[b]{1.0\linewidth}
  \centerline{\includegraphics[width=85mm]{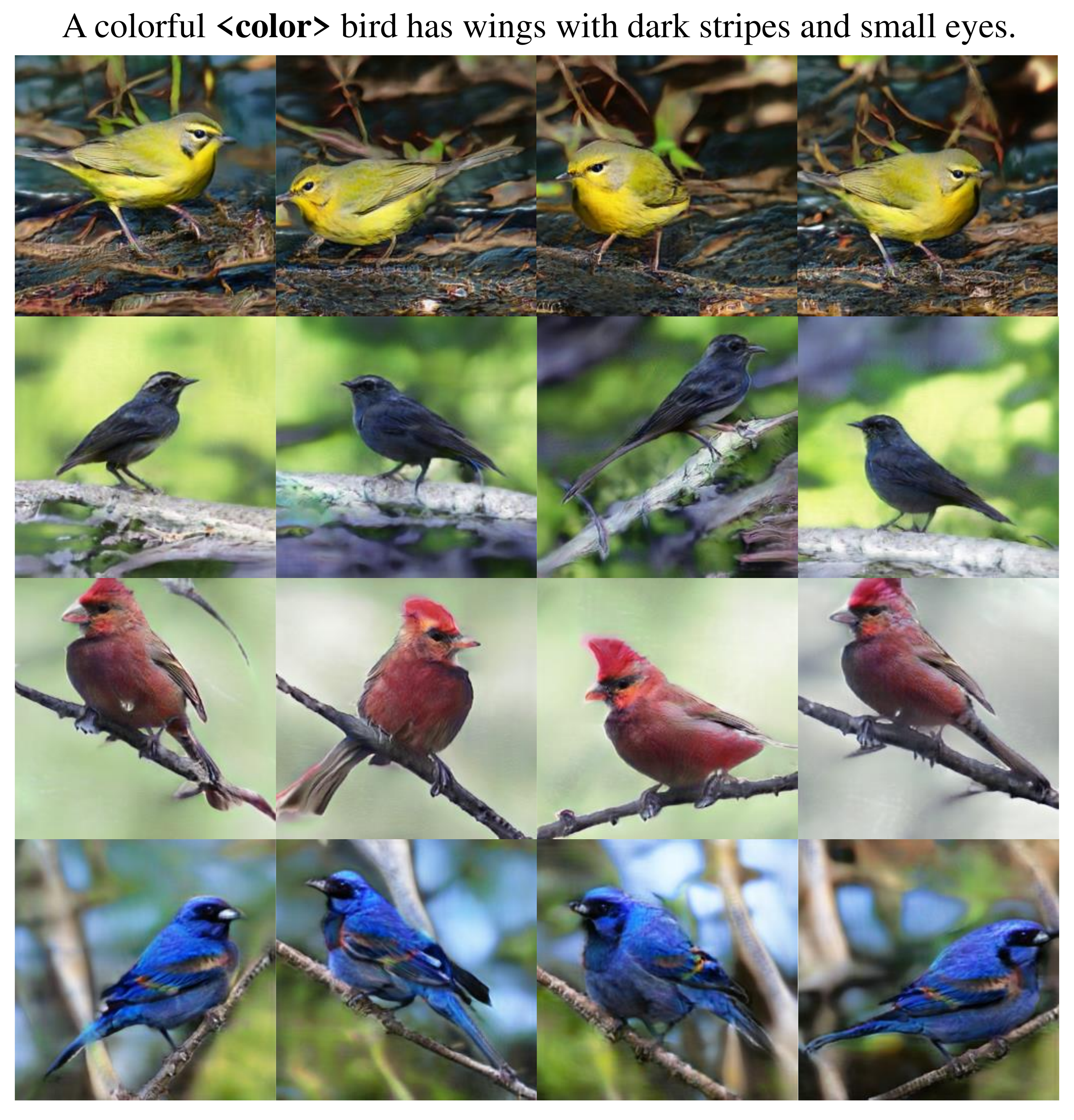}}
  \end{minipage}
  \caption{Generated images of the DTGAN by changing the color attribute value in the input text description, for four random draws.}
  \vspace{-0.1in}
  \label{fig7} %% label for entire figure
\end{figure}

\raggedbottom
Furthermore, to validate the sensitivity and diversity of our DTGAN, we generate birds by modifying just one word in the given text description. As can be seen in Figure~\ref{fig7}, the generated birds are similar but have different poses and shape for the same description. When we change the color attributes in the natural-language descriptions, the proposed DTGAN further produces semantically consistent birds according to the modified text. It means that our approach has the ability to accurately capture the modified part of the text description and to synthesize diverse images for the same natural-language text.
\subsection{Component Analysis}
In this section, we perform an extensive ablation study on the CUB dataset, so as to evaluate the contributions from different components of our DTGAN. The novel components in our model include a channel-aware attention module (CAM), a pixel-aware attention module (PAM), CAdaILN and a new type of visual loss (VL). We first quantitatively explore the effectiveness of each component by removing the corresponding part in the DTGAN step by step, i.e., 1) DTGAN, 2) DTGAN without the VL, 3) DTGAN without CAdaILN, 4) DTGAN without the PAM, 5) DTGAN without the CAM, 6) DTGAN without the CAM and PAM. All the results are reported in Table~\ref{tab:3}.
\begin{table}[t]
\begin{center}
\scalebox{.86}{
\begin{tabular}{c c c c c c c}
\hline
\multirow{2}*{ID} & \multicolumn{4}{c}{Components} &  \multicolumn{1}{c}{\multirow{2}*{IS $\uparrow$}} & \multicolumn{1}{c}{\multirow{2}*{FID $\downarrow$}} \\ 
\cline{2-5}
 &  CAM  &  PAM  &  CAdaILN  &  VL & & \\  
\hline
1 &  $\checkmark$ & $\checkmark$ & $\checkmark$ & $\checkmark$   &  $\bm{4.88\pm0.03}$ &  $\bm{16.35}$  \\ 

2 &  $\checkmark$ & $\checkmark$ & $\checkmark$ & - & $4.72\pm0.04$ &  19.23 \\

3 &  $\checkmark$ & $\checkmark$ & - & $\checkmark$   & $2.26\pm0.02$  &  91.53 \\

4 &  $\checkmark$ & - & $\checkmark$ & $\checkmark$   &  $4.71\pm0.05$ &  21.69 \\

5 &  - & $\checkmark$ & $\checkmark$ & $\checkmark$   & $4.60\pm0.07$  & 22.95 \\

6 &  - & - & $\checkmark$ & $\checkmark$   &  $4.54\pm0.04$ &  23.72 \\
\hline
\end{tabular}}
\end{center}
\caption{Ablation study of our DTGAN. CAM, PAM and VL represent the channel-aware attention module, the pixel-aware attention module and the visual loss, respectively. The best results are in bold.}
\label{tab:3}
\vspace{-0.1in}
\end{table}

By comparing Model 1 (DTGAN) with Model 2 (removing the VL), the VL significantly improves the IS from 4.72 to 4.88 and reduces the FID by 2.88 on the CUB dataset, which demonstrates the importance of adopting VL in the DTGAN. By exploiting CAdaILN in our DTGAN, Model 1 performs better than Model 3 (removing CAdaILN) on the IS and FID by 2.62 and 75.18, confirming the effectiveness of the proposed CAdaILN. Both Model 4 (removing the CAM) and Model 5 (removing the PAM) outperform Model 6 (removing the CAM and PAM), indicating that these two new types of attention modules can help the generator produce more realistic images. Furthermore, Model 1 achieves better results than both Model 4 and Model 5, which shows the advantage of combining the CAM and PAM.
% By comparing Model 1 (DTGAN) with Model 2 (removing the VL), the VL significantly improves the IS from 4.72 to 4.88 and reduces the FID by 2.88 on the CUB dataset, which demonstrate the importance of adopting the VL in the DTGAN. By exploiting CAdaILN in our DTGAN, Model 1 performs better than Model 3 (removing CAdaILN) on the IS and FID by 2.62 and 75.18, confirming the effectiveness of the proposed CAdaILN. Model 4 (removing the CAM) and Model 5 (removing the PAM) both outperform Model 6 (removing the CAM and PAM), indicating that these two new types of attention modules can help the generator produce more realistic images. Furthermore, Model 1 achieves better results than Model 4 and Model 5, which shows the advantage of combining the proposed CAM and PAM.
% \begin{figure*}[t]
%   \begin{minipage}[b]{1.0\linewidth}
%   \centerline{\includegraphics[width=180mm]{img8.pdf}}
%   \end{minipage}
%   \caption{Visualization of the channel-aware (detailed features) and pixel-aware (global shape) attention maps.}
%   \vspace{-0.1in}
%   \label{fig8} %% label for entire figure
% \end{figure*}
\begin{figure}[t]
  \begin{minipage}[b]{1.0\linewidth}
  \centerline{\includegraphics[width=85mm]{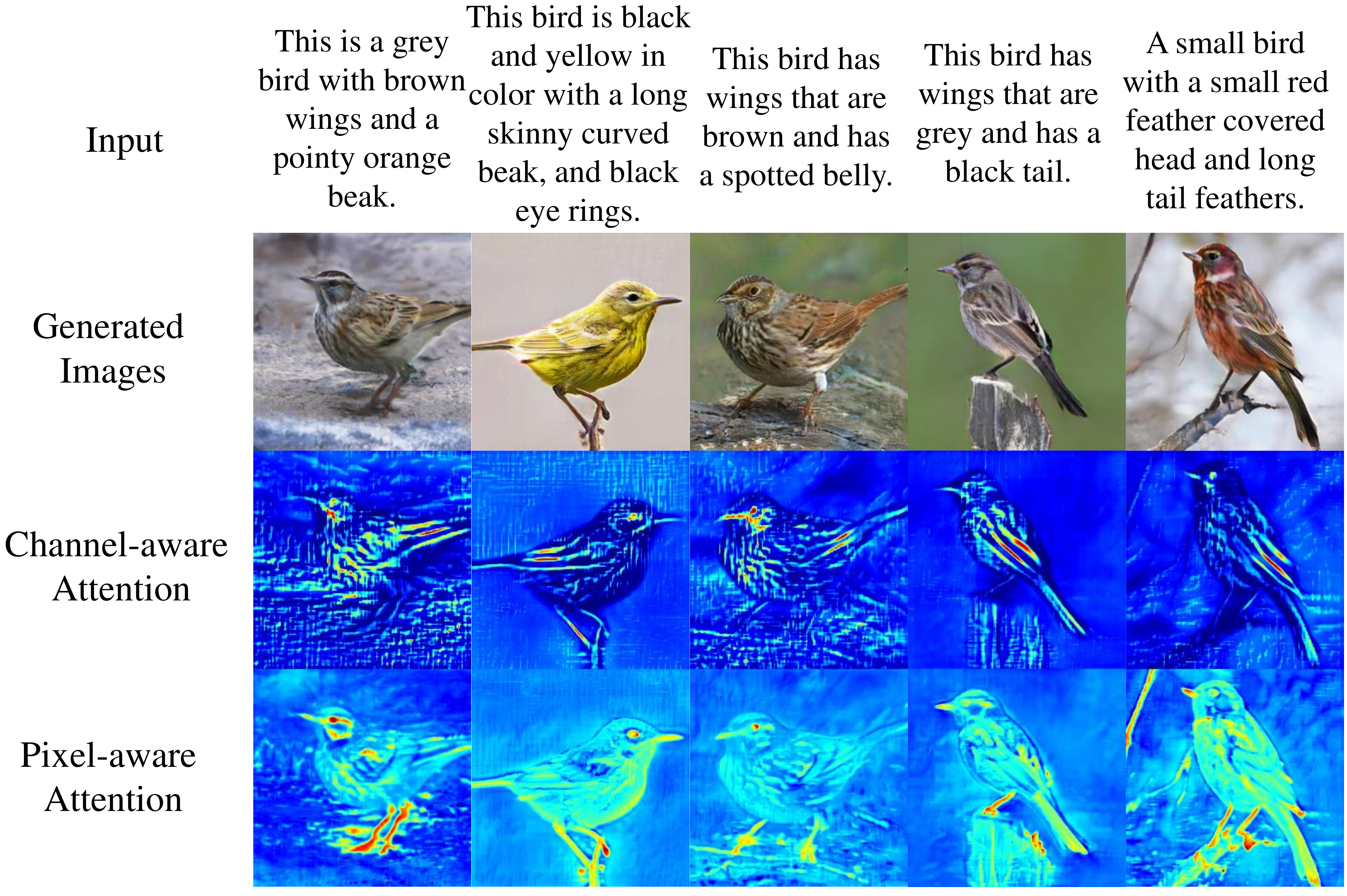}}
  \end{minipage}
  \caption{Visualization of the channel-aware (detailed features) and pixel-aware (global shape) attention maps.}
  \vspace{-0.1in}
  \label{fig8} %% label for entire figure
\end{figure}
\begin{figure*}[t]
  \begin{minipage}[b]{1.0\linewidth}
  \centerline{\includegraphics[width=180mm]{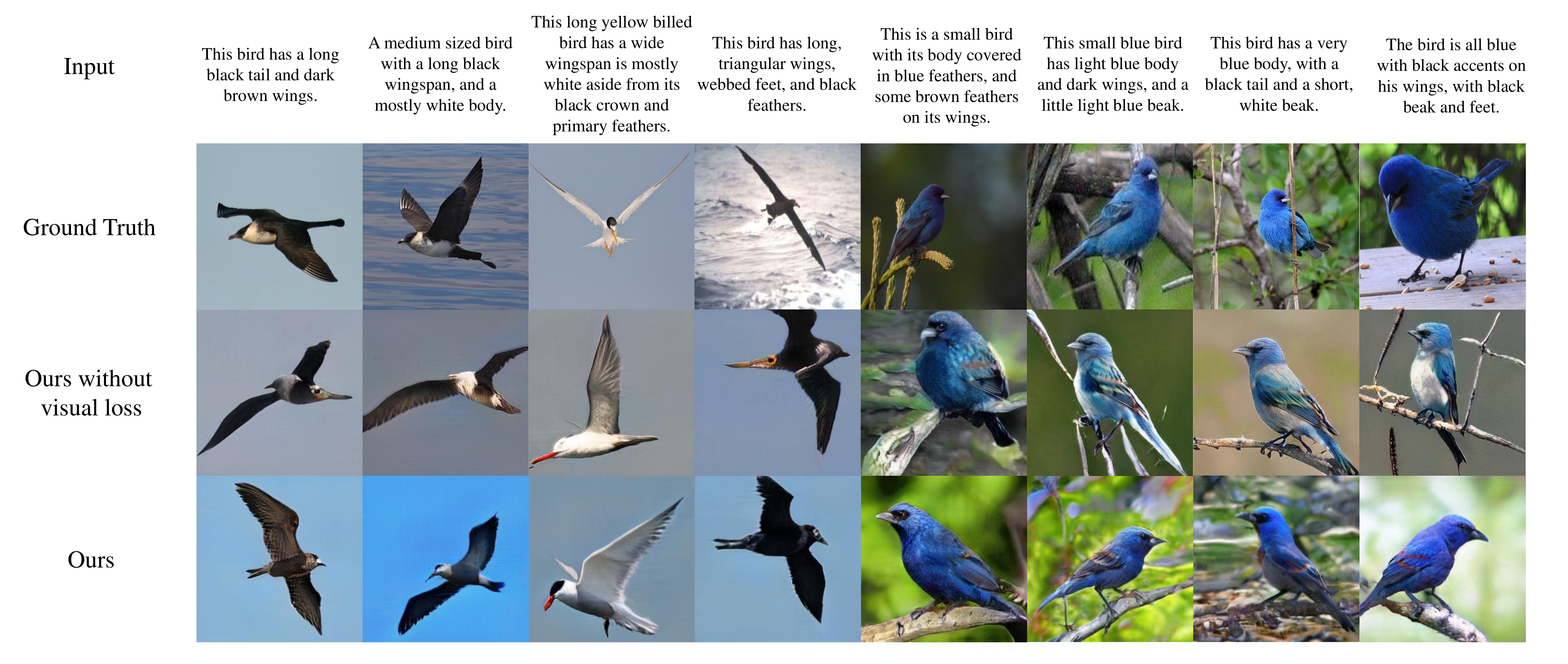}}
  \end{minipage}
  \caption{Visual comparison of the effect of our visual loss (VL) module, yielding more vivid shape and richer color distributions (bottom row).}
  \vspace{-0.1in}
  \label{fig9} %% label for entire figure
\end{figure*}

To better understand what has been learned by the CAM and PAM during training, we visualize the channel-aware and pixel-aware attention maps for different images in Figure~\ref{fig8}. We can see that in the  $2^{th}$ row, the eyes, beaks, legs and wings of birds are highlighted by the channel-aware attention maps. Meanwhile, in the $3^{th}$ row, the pixel-aware attention maps highlight most important areas of images, including the branches and the whole bodies of birds. This suggests that the CAM helps the generator focus on the crucial parts of birds, while the PAM guides the generator to refine the globally visual contents. Then, the generator can fine-tune the discriminative regions of images obtained by our attention modules.

\noindent\textbf{Visual Loss.} To balance the trade-off between image quality and semantic consistency, we investigate the hyper-parameter $\lambda_{1}$ by changing its value in the objective function. We test the value of $\lambda_{1}$ among 0.05. 0.10, 0.15, 0.20 and 0.30. The results are listed in Table~\ref{tab:4}. We can observe that the best performance is achieved on the CUB dataset if $\lambda_{1}$ is set to 0.1. Therefore, we use $\lambda_{1}$ as 0.1 in the  experiments.
% \begin{figure}[t]
%   \begin{minipage}[b]{1.0\linewidth}
%   \centerline{\includegraphics[width=85mm]{img09.pdf}}
%   \end{minipage}
%   \caption{Visual comparison of the effect of our visual loss (VL) module, yielding more vivid shape and richer color distributions (bottom row).}
%   \vspace{-0.1in}
%   \label{fig9} %% label for entire figure
% \end{figure}
\begin{table}
\begin{center}
\begin{tabular}{c c c c c}
\hline
Parameter & Values & IS $\uparrow$ & FID $\downarrow$\\
\hline
\multirow{5}*{$\lambda_{1}$} 
 &  0.05 & $4.74\pm0.05$ & 18.15 \\

 & 0.10 & $\bm{4.88\pm0.03}$ & $\bm{16.35}$ \\

 & 0.15 & $4.82\pm0.06$ & 16.75 \\

 &  0.20 & $4.59\pm0.04$ & 20.91 \\

 &  0.30 & $4.70\pm0.06$ & 20.28 \\
\hline
\end{tabular}
\end{center}
\caption{Evaluation of the DTGAN for different values of $\lambda_{1}$ which is the weight of the visual loss (VL) in the generator. The best result is in bold.}
\label{tab:4}
\vspace{-0.1in}
\end{table}

In addition, we conduct an ablation study to validate the effectiveness of the VL. The visual comparison between the DTGAN and our model without the VL is shown in Figure~\ref{fig9}. We can see that,  in the first two columns, the DTGAN without the VL fails to generate long-wingspan birds with reasonable shape and vivid wings. In the meantime, the proposed model without the VL synthesizes the blue birds which have rough color distributions and lack colorful details in the last two columns. However, the DTGAN produces realistic long-wingspan birds which have semantically consistent shape and colors, while also yielding blue birds with more vivid details and richer color distributions. This indicates that the VL has the ability to potentially ensure the quality of the generated image, including the shape and color distributions of objects in an image.
\begin{table}
\begin{center}
\begin{tabular}{c c c c c}
\hline
ID & Architecture & IS $\uparrow$ & FID $\downarrow$\\
\hline
1 &  Baseline & $2.26\pm0.02$ & 91.53 \\

2 &  +BN-sent & $4.67\pm0.07$ & 19.76 \\

3 &  +BN-word & $4.68\pm0.04$ & 19.46 \\

4 &  +CAdaILN & $\bm{4.88\pm0.03}$ & $\bm{16.35}$ \\

5 &  +CAdaILN-word & $4.71\pm0.07$ & 19.08 \\
\hline
\end{tabular}
\end{center}
\caption{Ablation study on CAdaILN. BN-sent indicates Batch Normalization conditioned on the global sentence vector, BN-word indicates Batch Normalization conditioned on the word vectors and CAdaILN-word indicates the CAdaILN function based on the word vectors.}
\label{tab:5}
\vspace{-0.1in}
\end{table}

\noindent\textbf{CAdaILN.} To further verify the benefits of CAdaILN, we conduct an ablation study for normalization functions. We first design a baseline model by removing CAdaILN from the DTGAN. Then we compare the variants of normalization layers. Note that BN conditioned on the global sentence vector (BN-sent) and BN conditioned on the word vectors (BN-word) are based on the conditional normalization methods in SDGAN \cite{yin2019semantics}, and CAdaILN based on the word vectors (CAdaILN-word) is revised on the basis of CAdaILN according to the word-level normalization method in SDGAN. The results of the ablation study are shown in Table~\ref{tab:5}. It can be observed that by comparing Model 2 with Model 4 and Model 3 with Model 5, CAdaILN significantly outperforms the BN layer whether using the sentence-level cues or the word-level cues. Moreover, by comparing Model 4 with Model 5, CAdaILN with the global sentence vector performs better than CAdaILN-word by improving the IS from 4.71 to 4.88 and reducing the FID from 19.08 to 16.35 on the CUB dataset, since sentence-level features are easier to be trained in our generator network than word-level features. The above analysis demonstrates the effectiveness of our designed CAdaILN.

%% file: Section/conclusion.tex
\section{Conclusion}
In this paper, we propose the Dual Attention Generative Adversarial Network (DTGAN), a novel framework for text-to-image generation, to generate high-quality realistic images which semantically align with given text descriptions, only employing a single generator/discriminator pair. DTGAN exploits two new types of attention modules: a channel-aware attention module and a pixel-aware attention module, to guide the generator to focus more on the text-relevant channels and pixels. In addition, to flexibly control the amount of change in shape and texture, Conditional Adaptive Instance-Layer Normalization (CAdaILN) is adopted as a complement to the attention modules. To further enhance the quality of generated images, we design a new type of visual loss which computes the L1 loss between the features of generated images and real images. DTGAN surpasses state-of-the-art results on both CUB and COCO datasets, which confirms the superiority of our proposed method. However, the improved visual quality comes with an apparent reduction in variation of generated images. Future work will be directed at mitigating this phenomenon by using larger training sets.